\documentclass[journal]{IEEEtran}
\usepackage{url}
\usepackage{xcolor}

\usepackage{graphicx}
\usepackage{amsmath}
\usepackage{amssymb}
\usepackage{array}
\usepackage{amssymb}
\usepackage{mathtext}
\usepackage{graphbox} 
\usepackage{multirow}
\usepackage{booktabs}
\usepackage[caption=false]{subfig}


\usepackage{mathtools}

\definecolor{newcolor}{rgb}{.8,.349,.1}
\begin{document}

\title{A Zero-Shot Sketch-based Inter-Modal Object Retrieval Scheme for Remote Sensing Images}

\author{Ushasi~Chaudhuri,~\IEEEmembership{Member,~IEEE,}
        Biplab~Banerjee,~\IEEEmembership{Member,~IEEE,}
        Avik~Bhattacharya,~\IEEEmembership{Senior Member,~IEEE,}
        and~Mihai~Datcu,~\IEEEmembership{Fellow,~IEEE}
\thanks{U.\ Chaudhuri, B.\ Banerjee and A.\ Bhattacharya are with the          Centre of Studies in Resources Engineering (CSRE), Indian Institute of Technology Bombay, India. Email: \{ushasi,bbanerjee\}@iitb.ac.in, avikb@csre.iitb.ac.in}
    \thanks{M. \ Datcu is with the German Aerospace Center (DLR), Germany. Email: mihai.datcu@dlr.de}
    }

\markboth{In submission to IEEE Geoscience and Remote Sensing Letters (GRSL)}%
{U.\ Chaudhuri et al.\MakeLowercase{\textit{et al.}}: Zero-shot Sketch-based inter modal}

\maketitle

\begin{abstract}
Conventional existing retrieval methods in remote sensing (RS) are often based on a uni-modal data retrieval framework. In this work, we propose a novel inter-modal triplet-based zero-shot retrieval scheme utilizing a sketch-based representation of RS data. The proposed scheme performs efficiently even when the sketch representations are marginally prototypical of the image. We conducted experiments on a new bi-modal image-sketch dataset called {\em Earth on Canvas} (EoC) conceived during this study. We perform a thorough bench-marking of this dataset and demonstrate that the proposed network outperforms other state-of-the-art methods for zero-shot sketch-based retrieval framework in remote sensing.
\end{abstract}


\IEEEpeerreviewmaketitle

\section{Introduction}
\IEEEPARstart{W}{ith} the advancement in sensor technology, huge amounts of data are being collected from various satellites. Hence, the task of target-based data retrieval and acquisition has become exceedingly challenging. Existing satellites essentially scan a vast overlapping region of the Earth using various sensing techniques, like multi-spectral, hyper-spectral, Synthetic Aperture Radar (SAR), video and compressed sensing, to name a few. With increasing complexity and different sensing techniques at our disposal, it has become our primary interest to design efficient algorithms to retrieve data from multiple data modalities, given the complementary information that are captured by different sensors. This type of problem is referred to as inter-modal data retrieval.

In remote sensing (RS), there are primarily two important types of problems, i.e., land-cover classification and object detection. In this work, we focus on target-based object retrieval part, which falls under the realm of object detection in RS. Object retrieval essentially requires high-resolution imagery for objects to be distinctly visible in the image. The main challenge with conventional retrieval approach using large scale databases is that, quite often, we do not have any query image sample of the target class at our disposal. The target of interest solely exists as a perception to the user in the form of an imprecise sketch. In such situations where a photo query is absent, it can be immensely useful if we can promptly make a quick hand-made sketch of the target. Sketches are a highly symbolic and hieroglyphic representation of data. One can exploit the notion of this minimalistic representative of sketch queries for sketch-based image retrieval (SBIR) framework~\cite{xu2019mental}.

While dealing with satellite images, it is imperative to collect as many samples of images as possible for each object class for object recognition with a high success rate. However, in general, there exists a considerable number of classes for which we seldom have any training data samples. Therefore, for such classes, we can use the zero-shot learning (ZSL) strategy. The ZSL approach aims to solve a task without receiving any example of that task during the training phase. This makes the network capable of handling an \textit{unseen} class (new class) sample obtained during the inference phase upon deployment of the network.

\begin{figure*}
    \centering
    \includegraphics[width=0.8\linewidth]{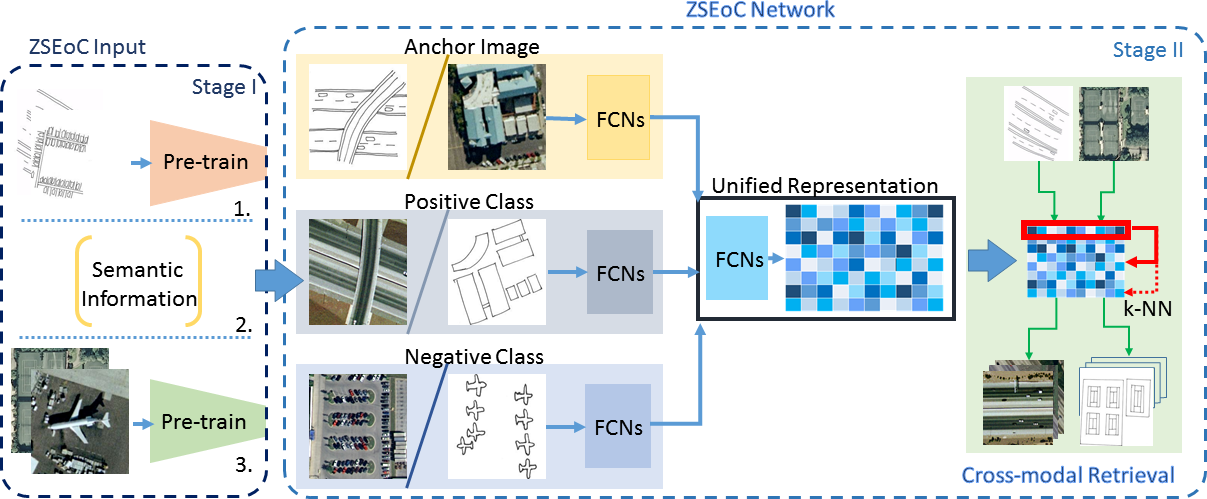}
    \caption{The complete pipeline of the ZSEoC framework for cross-modal retrieval of image$\rightarrow$sketch or sketch$\rightarrow$image.}
    \label{fig:block}
\end{figure*}

While SBIR has gained substantial attention in computer vision, this approach remains relatively unexplored in RS. A few notable attempts in this area include~\cite{kiran2018zero,shen2018zero,dira}, where the authors have utilized the standard bench-marked vision datasets of TU-Berlin and Sketchy. Lately, the idea of SBIR has been recognized as hugely relevant in RS of very high resolution (VHR) satellite images using deep features~\cite{xu2019mental,jiang2017sketch}. Both these tasks were accomplished using their proposed Aerial-SI dataset. However, this dataset remains unpublished for public usage. Further, Xu et al.,~\cite{Xu2020mental} exploited a related concept where they introduced the sketch-based remote sensing image retrieval (SBRSIR) dataset containing 20 classes, with 45 sketches and 200 images in each class. Even though the number of image samples is high, the number of sketch samples are comparatively low and inadequate for a conventional learning-based retrieval framework. 

Moreover, since sketches lack texture properties, additional samples were required for relevant discriminative learning. The authors in~\cite{Xu2020mental} proposed an adversarial technique for SBIR using a siamese metric learning technique. Adversarial techniques often lead to unstable training if the min-max problem is not intuitively designed. Also, they do not preserve the reverse embeddings of image-based sketch retrieval (IBSR) aspect of making the framework inter-modally-retrievable. The framework becomes inter-modally-retrievable if we can perform both SBIR and image-based sketch retrieval (IBSR). Although cross-modal retrieval in RS seems to be only partially explored~\cite{cmirnet}, the ZSL retrieval strategy for cross-modal sketch-image data remains practically unexplored, particularly in RS applications. 

In this work, we propose an efficient algorithm that performs a sketch-based inter-modal retrieval from RS images (figure~\ref{fig:block}). In this aspect, to the best of our knowledge, all the work that exists in literature only exploit the concept of sketch-based retrieval technique in RS. However, our proposed model is solely the one which introduces an inter-modal retrievable aspect associated with it. Furthermore, we also extend it for a ZSL based framework, which is a novel contribution in the field of RS. Even though we have proposed a framework for SBIR, the model is also robust to IBSR. We have carried all the experiments on an original sketch and high-resolution image bi-modal dataset called Earth on Canvas (EoC), proposed in this study. In the following section, we demonstrate the Zero-Shot retrieval architecture on the EoC dataset (ZSEoC)\footnote{{More details are provided in the supplementary file.}}.

\section{Methodology}
In the ZSL framework, we keep distinct classes in the training and the testing phase. The classes that are trained are referred to as the {\it seen classes}, while the classes that the network encounters during the testing phase are referred to as the {\it unseen classes}. For the unseen classes, the ZSL frameworks necessitate attribute information, which aids in recognizing the unseen classes during the testing phase. We refer to this attribute information as the {\it semantic information} in the remainder of the paper. 

Let us denote two streams of incoming data fields for Sketches and Images, denoted as $\mathcal{S}$ and $\mathcal{I}$, respectively. We aim to achieve an inter-modal retrieval model from $\mathcal{S}/\mathcal{I}$, given a query image from a different modality $\mathcal{I}/\mathcal{S}$. For the ZSL framework, we divide our dataset into seen and unseen class images for training and testing sets, respectively. If we use $L$ to define the set of labels and $\mathcal{W}$ as the semantic prototype, we can denote the training data as $\{\mathcal{S}^{tr}, \mathcal{I}^{tr}, L^{tr}, \mathcal{W}^{tr}\}$, and the test data as $\{\mathcal{S}^{ts}, \mathcal{I}^{ts}, L^{ts}, \mathcal{W}^{ts}\}$. Furthermore, we strictly ensure that the overlap between the seen and the unseen data is a null-set ($L^{tr} \cap L^{ts} = \varnothing$). With this structure, we design a unified latent feature representation for data from both the modalities. This approach allows us to achieve a zero-shot-based inter-modal retrieval framework, using the knowledge from the semantic information. The workflow of our proposed architecture is shown in figure~\ref{fig:block}.

\subsection{Network Construction}
The network is incorporated using a two-stage training strategy. In the first stage, we use transfer-learning from a network pre-trained on the Image-Net dataset~\cite{deng2009imagenet}. In the second stage, we design an encoder-decoder based architecture, in which, for the visual component, we use two separate encoders for the image and sketch data, while in the semantic component, we encode the attribute information. We use these representations to carry out visual-to-semantic mapping for the ZSL part. 

The visual encoders are a series of FCNs. Using feature embeddings from the pre-trained network, we stack four FCNs with 1024, 512, 256, and 128 nodes. In the semantic encoder part, we use a word-vector embedding for preserving the semantic topology. For example, runways and highways are more similar to each other than they are to parking lots; or a mobile home park is more similar to a building than it is to a baseball court. To accommodate these aspects, we use the standard {\tt word2vec} encoding of the semantic class labels~\cite{w2vmikolov2013distributed}. 

\subsection{Objective Function}
The overall objective function used in the proposed architecture is the sum of the following four loss functions described below:\\
\noindent\textbf{a) Cross-entropy (CE) loss ($\mathcal{L}_{ce}$):} An essential requirement while designing our model is to preserve the semantic class labels in the shared semantic space. For this purpose, we use the cross-entropy loss function for both the modalities to retain the semantic label information in the shared latent space~\eqref{eq:class},  
\begin{equation}\label{eq:class}
    \centering
    \mathcal{L}_{ce} =\text{CE} \left (w_s\mathcal{S}^{tr} \right ) +  \text{CE} \left (w_i\mathcal{I}^{tr}  \right )
\end{equation}
where $w_{s}$ and $w_{i}$ are the learnable parameters for creating the unified features. Here, $w_s\mathcal{S}= V_{s}$ and $w_i\mathcal{I}= V_{i}$, where $V_s$ and $V_i$ denote shared-space feature embeddings of the sketch and image instances, respectively. Even though we project the features onto a unified space, the shared features from both the modalities are different as each input is distinct. Hence, $V_s$ and $V_i$ denote the desired trained features of their corresponding modalities.

\noindent\textbf{b) Cross-triplet loss ($\mathcal{L}_{iii}$):} In the second stage, we employ three branches for data input streams, which we use to create the cross-triplets. Here, we use two types of triplets~\cite{chechik2010large}: a) we chose an anchor from the image dataset of class $c$, we select positive and negative samples from sketch data of class $c$ and any class other than $c$ (essentially, $c'$) respectively, b) we chose an anchor from the sketch data of class $c$, while we select positive and negative samples from image data of class $c$ and $c'$, respectively. This procedure displaces the negative class instances away from the anchor class by at least a margin $\alpha$ while making the same class instances of both the modalities closer. Effectively, it aids in decreasing the inter-modal distance while increasing the inter-class separability~\eqref{eq:3lt},
\begin{equation}\label{eq:3lt}
    \resizebox{1\hsize}{!}{  $ \mathcal{L}_{3a} =\operatorname{max} \left (
 { d\left(w_{s}  \mathcal{S}^{tr}_c  , w_{i} \mathcal{I}^{tr}_c\right)}
 - {d\left( w_s \mathcal{S}^{tr}_c ,w_i {\mathcal{{I}}^{tr}_{{c'}}}\right)}
 + \alpha, 0 \right )$}
\end{equation}
\begin{equation}\label{eq:3lt}
    \resizebox{1\hsize}{!}{  $ \mathcal{L}_{3b} =\operatorname{max} \left (
 { d\left(w_{i}  \mathcal{I}^{tr}_c  , w_{s} \mathcal{S}^{tr}_c\right)}
 - {d\left( w_i \mathcal{I}^{tr}_c ,w_s {\mathcal{{S}}^{tr}_{{c'}}}\right)}
 + \alpha, 0 \right )$}
\end{equation}
where $\alpha$ is a heuristically chosen margin value to push apart non-similar classes in the feature space. The sketch-anchored loss, $\mathcal{L}_{3a}$ and the image-anchored loss, $\mathcal{L}_{3b}$ together constitute the total cross-triplet loss, $\mathcal{L}_{iii}$.

\noindent\textbf{c) Cross-sample decoder loss ($\mathcal{L}_{dl}$):} The purpose of using this loss function is to make the unified latent space domain-independent. To achieve this, we bring the shared features of the two modalities closer to each other in the embedding space by performing an inter-modal data instance reconstruction. This loss helps in better class-wise alignment of both the modalities of data~\eqref{eq:dl}.  
\begin{equation}\label{eq:dl}
 \mathcal{L}_{dl} =||w_i^d V_s - w_i\mathcal{I}^{tr}_c||_{\text{F}}^2 +  ||w_s^d V_i - w_s\mathcal{S}^{tr}_c||_{\text{F}}^2
\end{equation}
where, $w^d_s$ and $w^d_i$ are the learnable parameters for the cross-sample decoder network. Here, we want $w^d_iV_s$ to learn its corresponding class feature encoding from $\mathcal{S}$ and similarly $w^d_s V_i $ from $\mathcal{I}$ (decoder network).

\noindent\textbf{d) Cross-projection loss ($\mathcal{L}_{cpl}$):} To make the data projections from both the modalities closer in the feature space, we minimize the mean-square difference between these two representations and the semantic information in the embedding space. This approach offers representations from both modalities akin to the semantic projection while bringing them closer to each other~\eqref{eq:cpl}.
\begin{equation}\label{eq:cpl}
    \centering
    \mathcal{L}_{cpl} =  ||w_i\mathcal{I}^{tr}_c - \mathcal{W}^{tr}_c||_{\text{F}}^2 + ||w_s\mathcal{S}^{tr}_c - \mathcal{W}^{tr}_c||_{\text{F}}^2 
\end{equation}

\noindent\textbf{Overall Objective function ($\mathcal{L}$):}
The final objective function is a combination of all the losses $\mathcal{L} = \mathcal{L}_{ce} + \mathcal{L}_{ iii} + \mathcal{L}_{dl} + \mathcal{L}_{cpl}$. For training this network, we perform an optimization on the final loss ($\mathcal{L}$). However, since we have a non-convex optimization problem at hand, we perform gradient descent on each of these losses individually while holding others constant. We solve the optimization problem by alternately minimizing each loss functions individually using the mini-batch gradient descent optimizer. Once we obtain the inter-modal embedding of each of the data sample, we can provide a query sample from either modality $\mathcal{S/I}$, and find the $k$-nearest neighbour ($k$-NN) instances from either of the modalities.

\subsection{The Earth on Canvas (EoC) Dataset}
We created the {\it Earth on Canvas} (EoC) dataset by utilizing a subset of image classes from the standard UC-Merced dataset~\cite{UCMerced2010}. The sketches were hand-drawn on paper by several amateur artists to avoid style-bias and were then photo-scanned with a resolution of 300-dpi. To place more attention to the sketch object from the sizeable background and increase its salience in the image, we cropped along the orthogonal hull of each sketch. To bring all the images to a similar dimension, we pad the remaining portion with a white background to produce a size of 256$\times$256-pixel dimension.

Out of the total 21 classes present in the Merced dataset, we drop the four land-cover classes (viz. agricultural, beach, chaparral, and forest), as we aim to solve an object retrieval problem.  Furthermore, we use a single class to represent the dense-residential, medium-residential, sparse-residential, and mobile-home park classes that describe the residential areas, as they have a visually similar appearance. Hence, the VHR-image data consists of 14 classes for which we have 100 image samples for each class (i.e., a total of 1400 optical images). We created 100 sketches for each of these 14 classes (i.e., a total of 1400 sketch images). Therefore, we have a total of 2800 images in the database containing both optical and sketch images sharing the same category labels. The classes in the dataset are Airplane, Baseball diamond, Buildings, Freeway, Golf course, Harbor, Intersection, Mobile home park, Overpass, Parking lot, River, Runway, Storage tanks, and Tennis court. 

\section{Experiments and Results}
We perform all our experiments on the proposed EoC dataset. In the ZSL experimental framework, we consider ten classes for training (i.e., seen classes) and four classes for testing (i.e., unseen classes). We use the last four classes, i.e., Runway, Water-tank, Tennis-court, and River classes as the unseen classes. For pre-training the network, we explore several standard models, namely: VggNet16~\cite{cheng2017}, ResNet50, and ResNet101~\cite{szegedy2017} ({results reported in Table~\ref{tab:comparison}}). By using transfer learning from these pre-trained networks, we perform fine-tuning on our dataset to encode the class labels in the extracted feature space. For the semantic vector, we use a 300-d vector embedding obtained from the pre-trained word2vec embedding procedure. We use two variants of the network: 1) we use the 300-d vector directly for training, and 2) we use two layers of FCNs to learn a 128-d distinct vector for semantic information.

In the encoder part of the visual streams, we design two variants of the network by taking two separate FCN networks for extracting a 300-d and a 128-d feature vector in the shared latent space representation. Both these networks have similar architecture. We have used four layers of FCNs with dimensions 1024, 512, 256, and eventually, 300 or 128 depending on the variant of the network. We have used batch-normalization and a leaky ReLU function to induce non-linearity. The network that learns the 300-d shared features is fed along with the 300-d {\tt word2vec} embeddings directly. We refer to this model as the fixed semantic vector variant. The network that learns 128-d feature vectors from the visual encoders uses a layer of FCN to project the 300-d semantic vector onto a 128-d feature space. Since semantic information is learnt from the network, we refer to this model as the latent semantic vector variant.
\begin{table}
\centering
\caption{SBIR performance of the proposed ZSEoC framework on the \textbf{EoC} dataset in terms of mAP (\%) and precision at top-100 (P@100) (\%) values.}
\begin{tabular}{lccm{4.6em}}
\toprule
    \multirow{1}{*}{\textbf{Task}} &
    \multicolumn{3}{c}{\textbf{EoC}}\\
    \cline{2-4}  
     & \textbf{mAP} & \textbf{P@100} &  \textbf{Feature dimension}\\
    \midrule
 Baseline-I (VggNet-16) & 0.221 & 0.234 & 4096 \\ 
 Baseline-II (ResNet-50) & 0.236 & 0.254 & 2048 \\ 
 Baseline-III (ResNet-101) & 0.269 & 0.284 & 2048 \\ 
 Baseline-IV (CNN) & 0.30 & 0.284 & 128 \\
 Baseline-V (Pre-train + CNN) &0.196& 0.284& 128 \\
 ZS-SBIR \cite{kiran2018zero} &0.395& 0.421&1024 \\ 
 ZSIH (binary) \cite{shen2018zero} & 0.452 & 0.487 & 64\\ \midrule
 \textbf{ZSEoC-300} (fixed semantic vector) & \textbf{0.686}&\textbf{{0.698}} & {300}\\ 
\textbf{ZSEoC-128} (latent semantic vector) & \textbf{{0.674}}&\textbf{0.732} & {128}\\ \bottomrule
\end{tabular}\label{tab:comparison} \vspace{-3mm}
\end{table}

We create 14,000 triplets for each type of anchor while training the network. In each batch, we make sure that an equal number of sketch-anchored triplets and image-anchored triplets are provided to avoid any training bias. To the best of our knowledge, there does not exist any ZSL based inter-modal retrieval algorithm in the literature. Therefore, in this respect, we perform a few baselines for the sake of comparison. In Baseline-I, we use the pre-trained weights from VggNet-16 and use a $k$-NN based approach to find the top-$k$ retrieved vales. For Baseline-II and Baseline-III, we use a similar framework, but with ResNet-50 and ResNet-101, respectively. For Baseline-IV, we obtain 128-d features by using a series of 2D convolution layers, directly from the images and sketches. We denote the convolutional layer parameters as “conv $<$receptive field size$>-<$number of channels$>$". We use two layers of conv3-64, followed by two layers of conv3-128. Both these pair of convolution layers are followed by a maxpool and a batch-normalization layer. This is then followed by another conv3-256 and maxpool layer. Finally, the output is fed to a fully-connected layer of dimension 128. Ultimately, we combine the ResNet-101 pre-trained network with three subsequent layers of 2D convolution (last three layers from Baseline-IV) and use it as Baseline-V. 

The proposed framework is also compared with the state-of-the-art (SOTA) ZSL sketch-based image retrieval network (ZS-SBIR)~\cite{kiran2018zero}. Similarly, we also compare our results with the Zero-shot Image Hashing (ZSIH) network~\cite{shen2018zero}. This technique is a zero-shot sketch image retrieval model wherein the network uses a generative hashing scheme for constructing the semantic information. It is noteworthy that these are solely SBIR type networks and do not support inter-modal retrieval tasks. Therefore, to maintain integrity in comparison, the SBIR method was evaluated using only the sketch-anchored triplets. It can be noted from Table~\ref{tab:comparison} that our ZSEoC framework outperforms the current SOTA methods. Table~\ref{tab:unicross} shows the inter-modal retrieval results, along with the uni-modal (sketch$\rightarrow$sketch \& images$\rightarrow$images) retrieval results. 

\begin{figure}
\centering
\subfloat[Image]{%
\includegraphics[width=0.47\columnwidth]{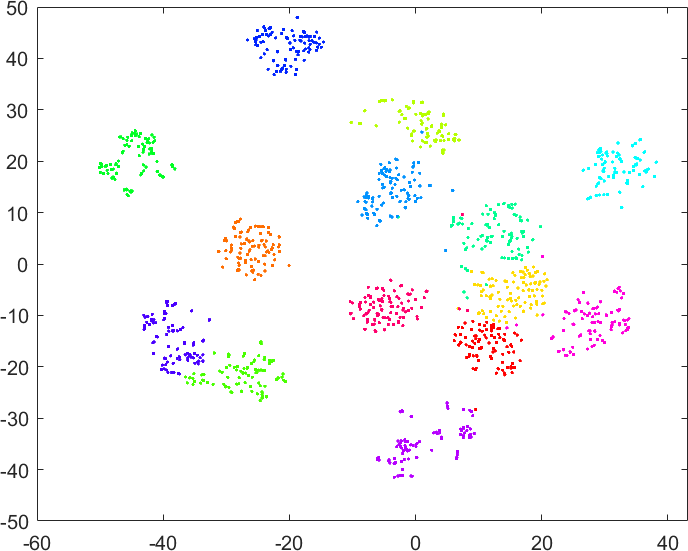}}
\hfill
\subfloat[Sketch]{%
\includegraphics[width=0.47\columnwidth]{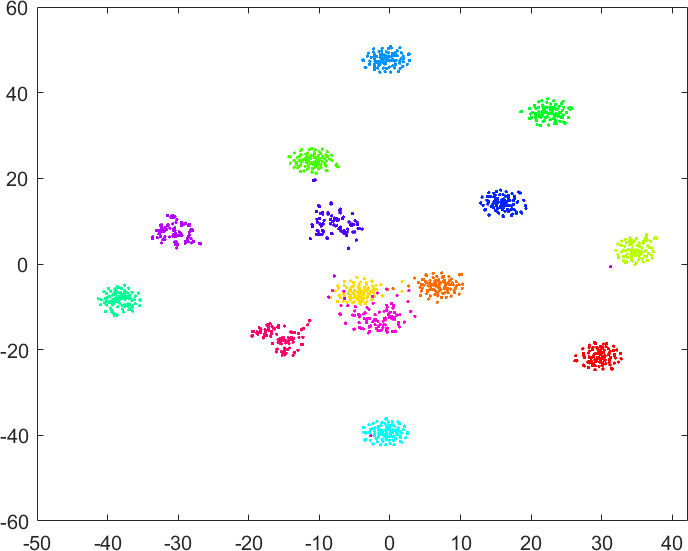}}
\caption{Two-dimensional scatter plots of high-dimensional features generated with $t$-SNE of image and sketch features, in the shared latent space, trained with a fixed-semantic vector. Clusters with distinct colours denote separate classes in the dataset.}
\label{fig:tsne}\vspace{-3mm}
\end{figure}

\begin{figure*}
   \centering 
   \includegraphics[height=1.45cm]{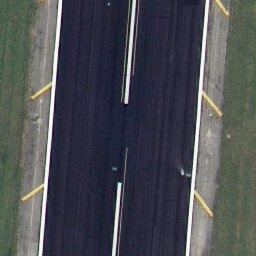} 
   \fcolorbox{green}{yellow}{\includegraphics[height=1.45cm]{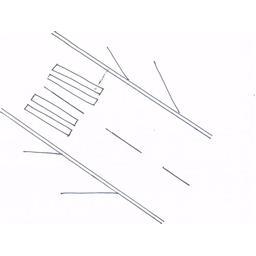}}
   \fcolorbox{green}{yellow}{\includegraphics[height=1.45cm]{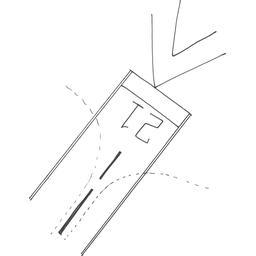}}
   \fcolorbox{green}{yellow}{\includegraphics[height=1.45cm]{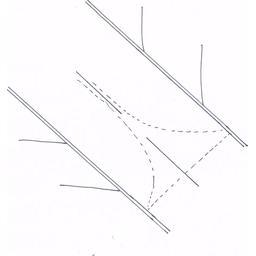}}
   \fcolorbox{green}{yellow}{\includegraphics[height=1.45cm]{images/55.jpg}}
    \fcolorbox{red}{yellow}{\includegraphics[height=1.45cm]{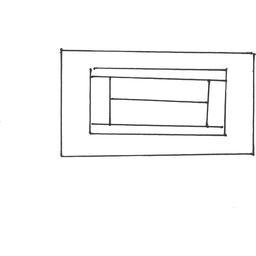}}
   \fcolorbox{green}{yellow}{\includegraphics[height=1.45cm]{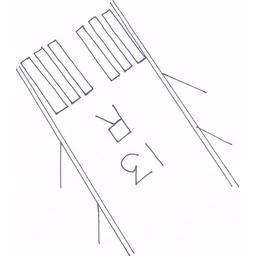}}
   \fcolorbox{green}{yellow}{\includegraphics[height=1.45cm]{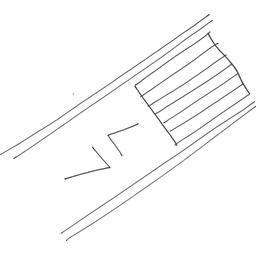}}
   \fcolorbox{green}{yellow}{\includegraphics[height=1.45cm]{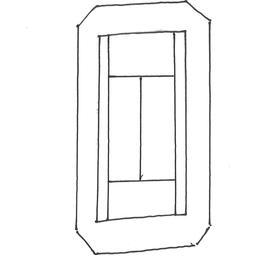}}
   \fcolorbox{green}{yellow}{\includegraphics[height=1.45cm]{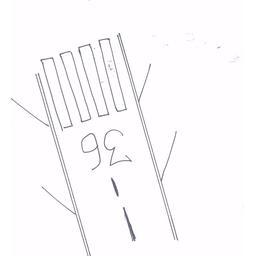}}\\
   {\includegraphics[height=1.45cm]{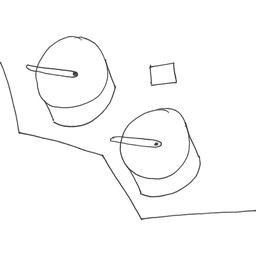}} 
   \fcolorbox{green}{yellow}{\includegraphics[height=1.45cm]{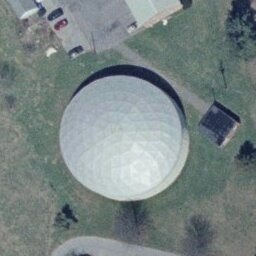}}
   \fcolorbox{red}{yellow}{\includegraphics[height=1.45cm]{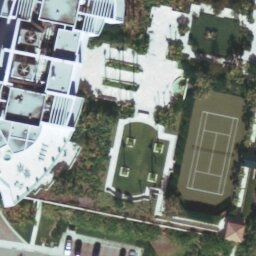}}
   \fcolorbox{green}{yellow}{\includegraphics[height=1.45cm]{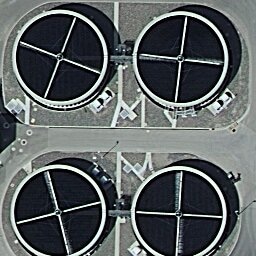}}
    \fcolorbox{green}{yellow}{\includegraphics[height=1.45cm]{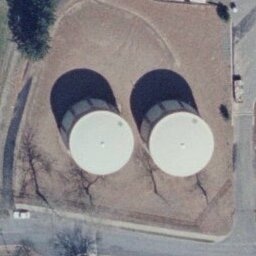}}
   \fcolorbox{green}{yellow}{\includegraphics[height=1.45cm]{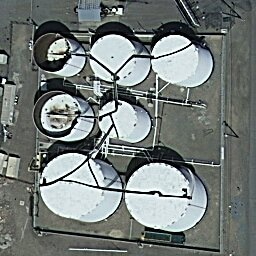}}
   \includegraphics[height=1.45cm]{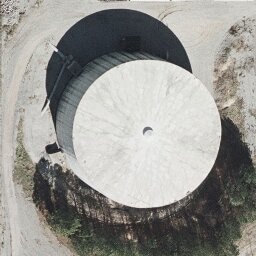}
   \fcolorbox{green}{yellow}{\includegraphics[height=1.45cm]{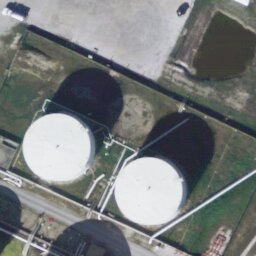}}
   \fcolorbox{green}{yellow}{\includegraphics[height=1.45cm]{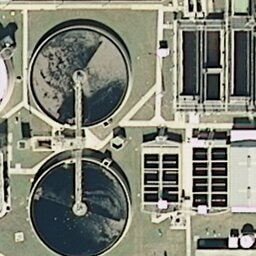}}
   \fcolorbox{green}{yellow}{\includegraphics[height=1.45cm]{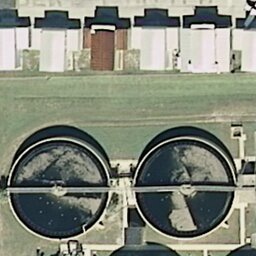}}\\
   {\includegraphics[height=1.45cm]{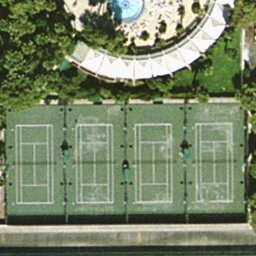}} 
   \fcolorbox{green}{yellow}{\includegraphics[height=1.45cm]{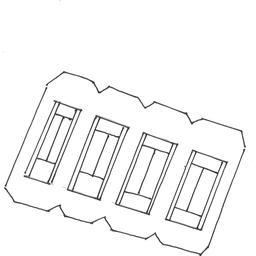}}
    \fcolorbox{green}{yellow}{\includegraphics[height=1.45cm]{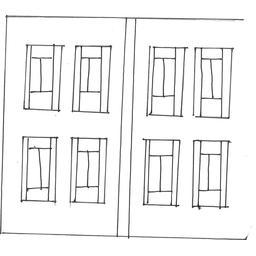}}
    \fcolorbox{green}{yellow}{\includegraphics[height=1.45cm]{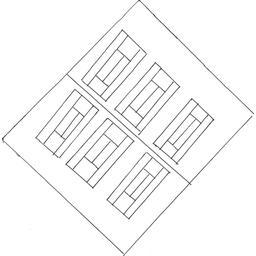}}
       \fcolorbox{green}{yellow}{\includegraphics[height=1.45cm]{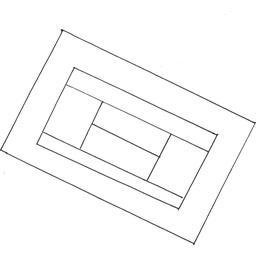}}
   \fcolorbox{red}{yellow}{\includegraphics[height=1.45cm]{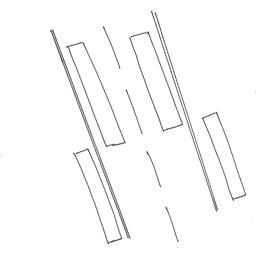}}
   \fcolorbox{green}{yellow}{\includegraphics[height=1.45cm]{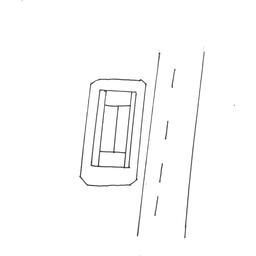}}
   \fcolorbox{green}{yellow}{\includegraphics[height=1.45cm]{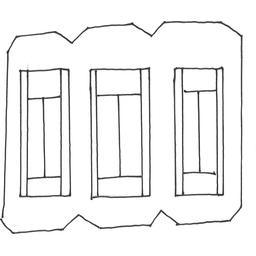}}
   \fcolorbox{red}{yellow}{\includegraphics[height=1.45cm]{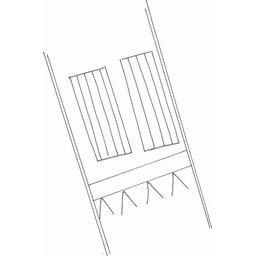}}
   \fcolorbox{green}{yellow}{\includegraphics[height=1.45cm]{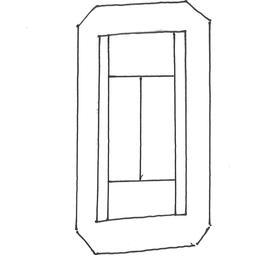}}\\
      {\includegraphics[height=1.45cm]{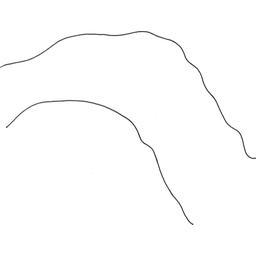}} 
   \fcolorbox{green}{yellow}{\includegraphics[height=1.5cm]{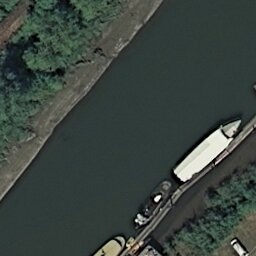}}
    \fcolorbox{green}{yellow}{\includegraphics[height=1.5cm]{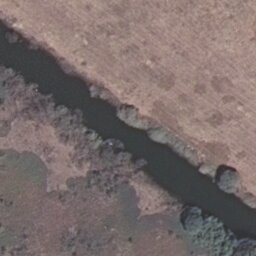}}
    \fcolorbox{green}{yellow}{\includegraphics[height=1.5cm]{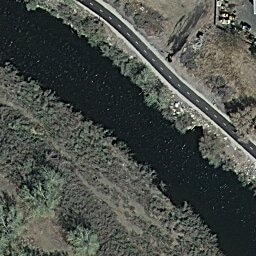}}
   \fcolorbox{green}{yellow}{\includegraphics[height=1.5cm]{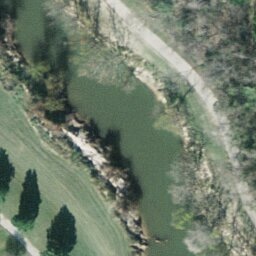}}
   \fcolorbox{green}{yellow}{\includegraphics[height=1.5cm]{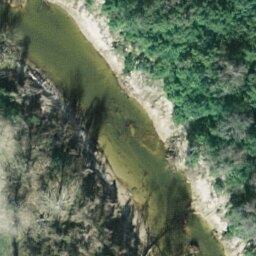}}
   \fcolorbox{green}{yellow}{\includegraphics[height=1.5cm]{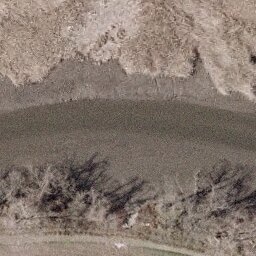}}
   \fcolorbox{green}{yellow}{\includegraphics[height=1.5cm]{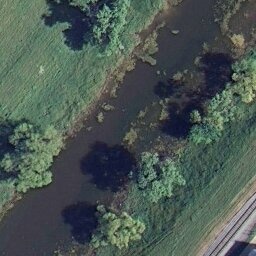}}
   \fcolorbox{green}{yellow}{\includegraphics[height=1.5cm]{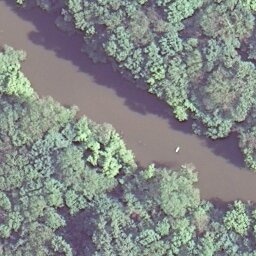}}
    \fcolorbox{green}{yellow}{\includegraphics[height=1.5cm]{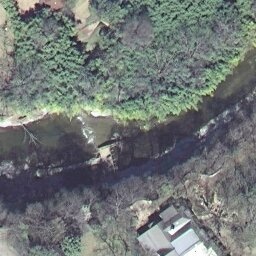}}
    \caption{A few top retrieved results from the zero-shot inter-modal framework. Alternate rows represent \textbf{Sketch$\rightarrow$Image} and \textbf{Image$\rightarrow$Sketch} retrievals.}
    \label{fig:retrieved}\vspace{-2mm}
\end{figure*}

It is interesting to note that the uni-modal retrieval results are likewise efficiently encoded in this unified feature embedding space along with the inter-modal ones, thus providing better retrieval outcomes than the inter-modal results. Therefore, the aforementioned indicates the efficiency of the shared embedding space. Furthermore, figure~\ref{fig:tsne} shows two-dimensional scatter plots of the high-dimensional features generated with the $t$-distributed stochastic neighbor embedding ($t$-SNE) algorithm for image and sketch features, in the shared latent space that is trained with a fixed-semantic vector. Moreover, it can be noticed that the data instances for both the modalities are separated and grouped in the unified space using the ZSEoC model. Here, we also present a few inter-modal retrieval results in figure~\ref{fig:retrieved}, where the images with green borders indicate the correctly retrieved images. In contrast, the ones with red borders show incorrect results. 
\begin{table}
\centering
\caption{Inter-modal retrieval performance of the proposed ZSEoC framework on the EoC dataset.}
\begin{tabular}{m{5.6em} ll m{5.6em} ll}
\toprule
    \multirow{1}{*}{\textbf{Inter-modal}} &
    \multicolumn{2}{c}{\textbf{EoC}} &
    \multirow{1}{*}{\textbf{Uni-modal}} &
    \multicolumn{2}{c}{\textbf{EoC}} \\
    \cline{2-3}  \cline{5-6}   
   & \textbf{mAP} & \textbf{P@100}& & \textbf{mAP} & \textbf{P@100}  \\
    \midrule
     Sketch$\rightarrow$Image & 0.686 & 0.698 &  Image$\rightarrow$Sketch  & 0.612 & 0.632 \\  
     Sketch$\rightarrow$Sketch  & 0.719 & 0.737 &
     Image$\rightarrow$Image   & 0.839 & 0.855 \\ 
      \bottomrule
\end{tabular}\label{tab:unicross} \vspace{-2mm}
\end{table}

\begin{figure}[h]
    \centering
    \includegraphics[width=0.35\textwidth]{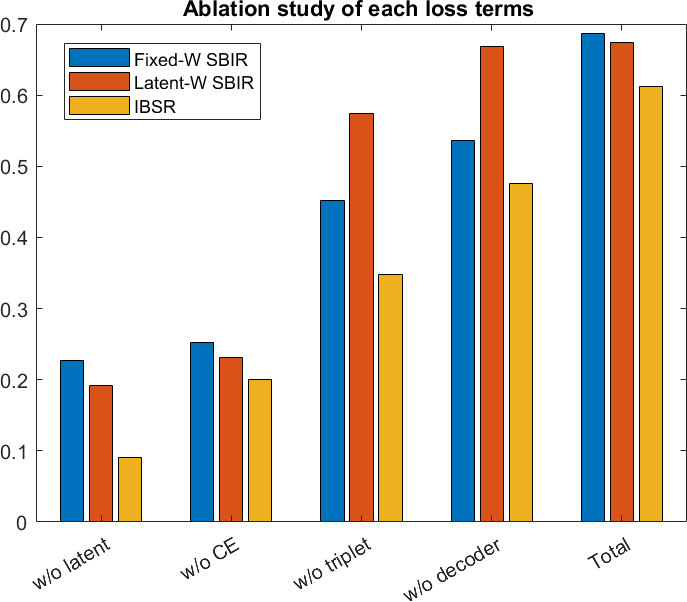}\vspace{-4mm}
    \caption{Ablation study for understanding the contribution of each loss term in the objective function. We study SBIR with fixed semantic vector and latent semantic vector, and IBSR with latent semantic vector.}
    \label{fig:ablation}
\end{figure}

\noindent \textbf{Ablation Studies:} To investigate the effect of each loss function, we perform an ablation study. For this purpose, we run our experiments on three problem sets: i) sketch-based image retrieval (SBIR) with fixed semantic space of 300-d, ii) SBIR with a latent-semantic space of 128-d, and iii) IBSR. For the first set of experiments, we used the total loss function, except for the latent loss, which primarily aids in bringing the two modalities closer to each other for a consistent retrieval purpose. However, we can notice from Fig~\ref{fig:ablation}, that without this loss function, there is a noteworthy decrease in the overall performance of the system. 

In the second set of experiments, we leave out the cross-entropy loss function. In doing so, the features in the shared embedding space lose their inter-class distances resulting in an ineffective retrieval, which can be seen in figure~\ref{fig:ablation}. In the third set of experiments, we exclude the cross-triplets loss function from the overall objective function. A significant variation in the performance between the SBIR and the IBSR modules is noticed from figure~\ref{fig:ablation}. Therefore it is noted that the cross-triplets loss aids in boosting both the inter-modal retrieval frameworks. However, we can observe that the retrieval ability of the framework significantly increases when retaining only the single cross-triplet (i.e., either sketch anchored or image anchored), while decreasing for the other. Therefore keeping both sketch and image anchored triplets leads to an optimum trade-off between both the performances. 

In the fourth set of experiments, we exclude the decoder loss function from $\mathcal{L}$. Surprisingly, we still observe the excellent performance of the framework. However, the inclusion of this loss function provides an additional impetus in the execution of the framework, making its performance better than the SOTA. The last set of bar graphs in figure~\ref{fig:ablation} displays the performance of the complete model with the total objective function. 
\vspace{-1mm}
\section{Conclusion}
We propose an aerial image and sketch-based inter-modal zero-shot learning framework in remote sensing application\footnote{The Earth on Canvas dataset and the codes developed in this work are made available at: \url{https://github.com/ushasi/Earth-on-Canvas-dataset-sample}.}. Our primary motivation is to project the multi-modal data into a shared space for inter-modal retrieval. We extend this concept to create a framework wherein we might not have any training samples for some classes; however, there is a possibility that we might find them at any given instance. We exploit the notion of sketch-based image retrieval to tackle the problem of insufficient query image for a target within a class during the retrieval process. We propose a novel zero-shot inter-modal architecture for RS image retrieval using the EoC dataset introduced in this work. The performance of the proposed algorithm exceeds the current SOTA results in SBIR.
\vspace{-2mm}

\bibliographystyle{IEEEtran}
\bibliography{mybib.bib}

\end{document}


\title{A Zero-Shot Sketch-based Inter-Modal Object Retrieval Scheme for Remote Sensing Images \\
\textbf{Supplementary Material}}

\author{U.~Chaudhuri, B.~Banerjee, A.~Bhattacharya, M.~Datcu}


\maketitle
\section{Cross-Triplet Loss Function}

The triplet loss function is a standard metric loss function~[1] defined as:

$\mathcal{L} \left ( a, p, n \right ) =\operatorname{max} \left (
 {\| f \left ( a \right ) - f \left ( p \right ) \|}^2
 - {\| f \left ( a \right ) - f \left ( n \right ) \|}^2
 + \alpha, 0 \right )$,
 
\noindent where, $a \in A$ is the anchor instance, $p \in P$ and $n \in N$ are the positive and negative instances respectively. $f \in \mathbb{R}^{d}$ is the embedding representation in $d$ dimensional Euclidean space.  

\begin{figure}[!h]
    \centering
    \includegraphics[width=0.95\columnwidth]{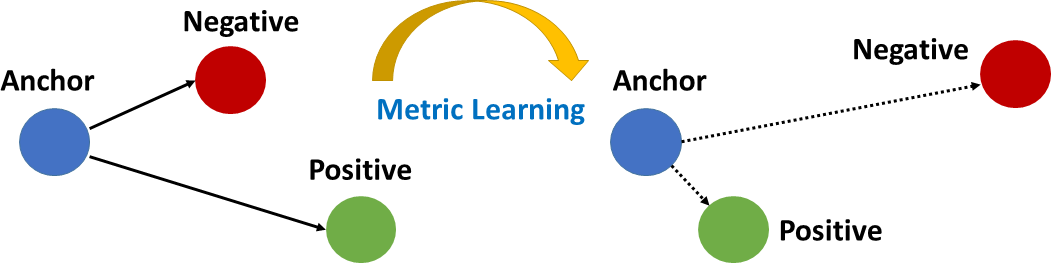}
    \caption{A schema displaying the basic principle of minimizing the triplet loss function [2]: increase the distance between the anchor and negative sample and decrease the distance between the anchor and positive sample in the embedding space.}
    \label{fig:metric}
\end{figure}

In the cross-modal image retrieval case, we consider $a \in A$ from one modality while $p \in P$ and $n \in N$ are considered from other modalities. The hyper-parameter, $\alpha$ in this loss function is called the margin. The margin defines the separation between dissimilarities, which in turn helps us better distinguish among two images. Hence, for this particular framework, we term this the cross-triplet loss function. Figure~\ref{fig:metric} shows the principle of this cross-triplet loss. Let us consider the two possible cases, keeping $\alpha = 1$, $d(a,p) = x$ and $d(a,n) = y$:
 
 \textbf{Case I:} If  $( (1+x) > y)$, this indicates that the distance between the negative and anchor instance is less than the distance between the anchor and positive instance plus a margin. This scenario is undesired as we want to have larger $d(a,n)$. Therefore the resultant loss is minimized so that the negative sample is driven beyond the margin.
 
 \textbf{Case II:} If  $( (1+x) \le y)$, this satisfies the condition that the distance between the anchor and the positive instance is smaller (plus margin) than the one with its negative sample. This scenario is what we desire, and hence, in this case, the maximum value is 0, implies that the resultant loss $\mathcal{L} = 0$. Thus, for this set of samples, we do not minimize the loss.
 
So the idea is, whenever we provide two images from same classes and different modalities, we are essentially reducing the inter-modal distance (because of the difference loss) and maximizing the inter-class distance (because of the margin).  

\begin{algorithm}[H]
\SetAlgoLined
\KwData{$ \{\mathcal{S}^{tr}, \mathcal{I}^{tr}, \mathcal{W}^{tr}, 
	{L}^{tr}\}$}
\KwResult{Obtain ${V_s}$ $\&$ ${V_i}$ }
Pre-train {for} dataset $\mathcal{S}$ and $\mathcal{I}$ using $L$ 

Construct $\mathcal{W}^{tr}$ using {\tt word2vec} 

\While{$\mathcal{L} {<} \epsilon$}{
Construct sketch and image-anchored cross triplets\;\\
$T_s \longleftarrow \{a_s^c,p_i^c,n_i^{c'} \}$; \, $a \in A$, $p \in P$, $n \in N$\;\\
$T_i\longleftarrow \{a_i^c,p_s^c,n_s^{c'}\}$ \;\\
   $\underset{w_i, w_s,w_{s}^d, w_{i}^d}{\min}\{ \mathcal{L}_{cpl} + \mathcal{L}_{iii} + \mathcal{L}_{ce} + \mathcal{L}_{dl} \}$\;
 }
 \KwResult{ $w_i, w_s,w_{s}^d, w_{i}^d$}
 ${V_s}$ $\&$ ${V_i} \longleftarrow w_i, w_s,w_{s}^d, w_{i}^d$\;
 
\vspace{5mm}
\textbf{Inference Stage}
 
$k$-NN (from $V_i$) using query embedding (from $V_s$)\;
\caption{ZSEoC}
\end{algorithm}

\noindent \underline{{\textbf{References:}}}
\\ \\
\noindent [1]. Gal Chechik, Varun Sharma, Uri Shalit, and Samy Bengio. "Large Scale Online Learning of Image Similarity Through Ranking." Journal of Machine Learning Research 11, no. 3 (2010).
\\ \\
\noindent [2]. Florian Schroff, Dmitry Kalenichenko, and James Philbin. "Facenet: A unified embedding for face recognition and clustering." Proceedings of the IEEE conference on computer vision and pattern recognition. 2015.

\newpage
\section{Sketch samples from Earth on Canvas dataset}
\begin{figure*}[!h]
\centering
\begin{tabular}{cccc}
\subfloat[Airplane]{{\includegraphics[width=0.2\linewidth]{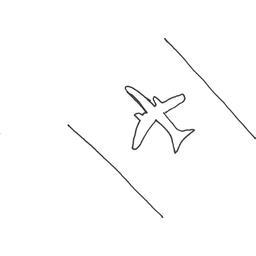}}} &
\subfloat[Baseball diamond court]{{\includegraphics[width=0.2\linewidth]{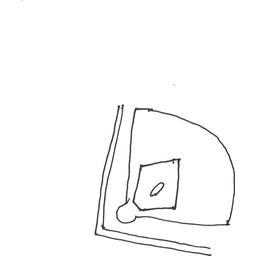}}} &
\subfloat[Buildings]{\includegraphics[width=0.2\linewidth]{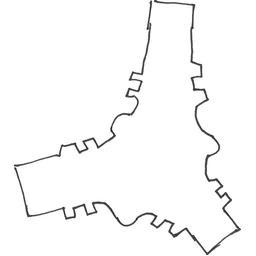}} &
\subfloat[Freeway]{\includegraphics[width=0.2\linewidth]{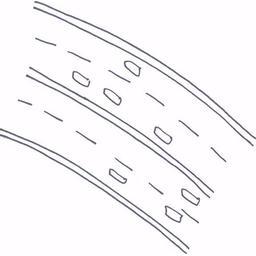}} \\
\subfloat[Golf course]{\includegraphics[width=0.2\linewidth]{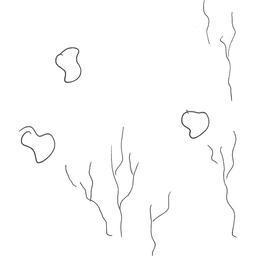}} &
\subfloat[Harbour]{\includegraphics[width=0.2\linewidth]{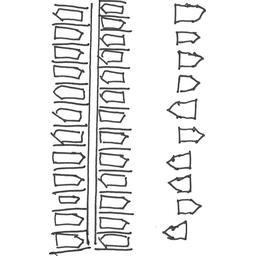}}&
\subfloat[Intersection]{\includegraphics[width=0.2\linewidth]{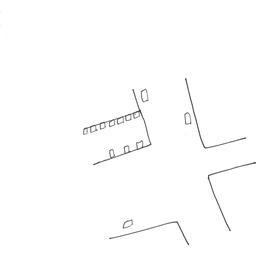}} &
\subfloat[Mobile home park]{\includegraphics[width=0.2\linewidth]{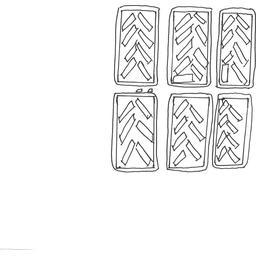}}\\
\subfloat[Overpass]{\includegraphics[width=0.2\linewidth]{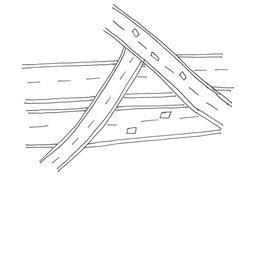}} &
\subfloat[Parking lot]{\includegraphics[width=0.2\linewidth]{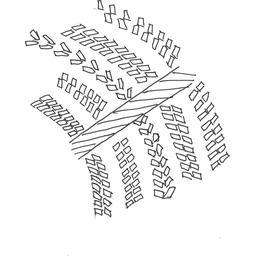}} &
\subfloat[River]{\includegraphics[width=0.2\linewidth]{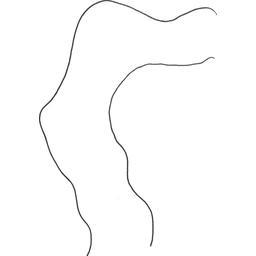}} &
 {\includegraphics[width=0.2\linewidth]{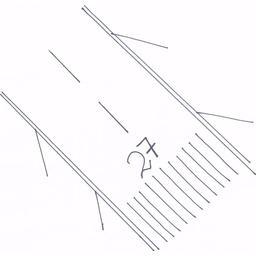}}\\
&
\subfloat[Storage tank]{\includegraphics[width=0.2\linewidth]{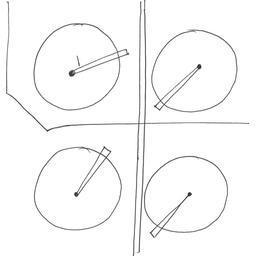}} &
\subfloat[Tennis court]{\includegraphics[width=0.2\linewidth]{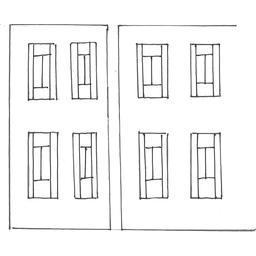}}
&\\
\end{tabular}
\caption{A few sample sketch instances of each class from the EoC dataset.}
\label{fig:sampleEoC}
\end{figure*}

\newpage
\section{Supplementary Experiments and Results}

\subsection{$t$-SNE plot of image and sketch data}
\begin{figure*}[!h]
\centering
\begin{tabular}{cc}
\subfloat[$t$-SNE of images before processing]{\includegraphics[width=0.47\linewidth]{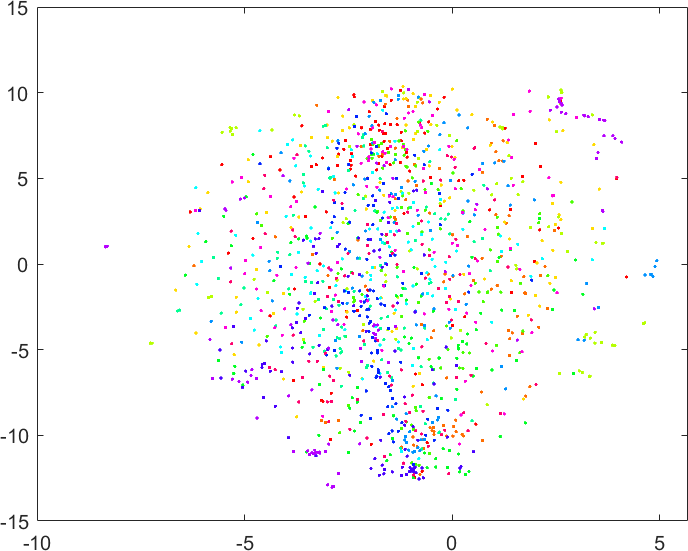}} &
\subfloat[$t$-SNE of sketches before processing]{\includegraphics[width=0.47\linewidth]{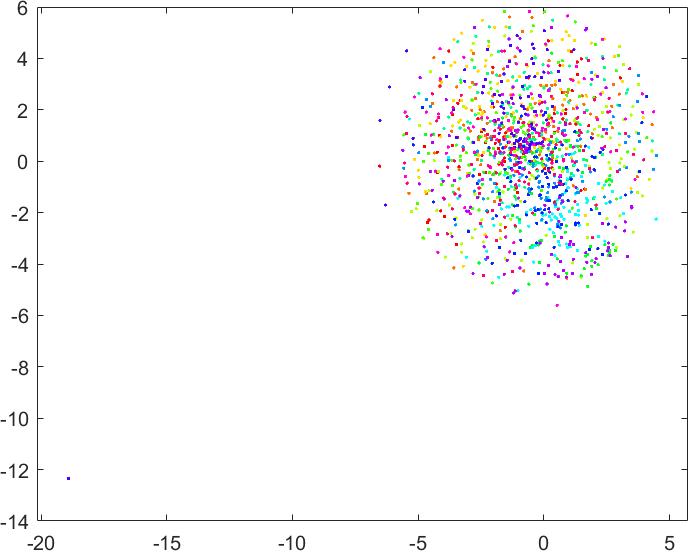}} \\
\subfloat[$t$-SNE of images in shared latent space]{\includegraphics[width=0.47\linewidth]{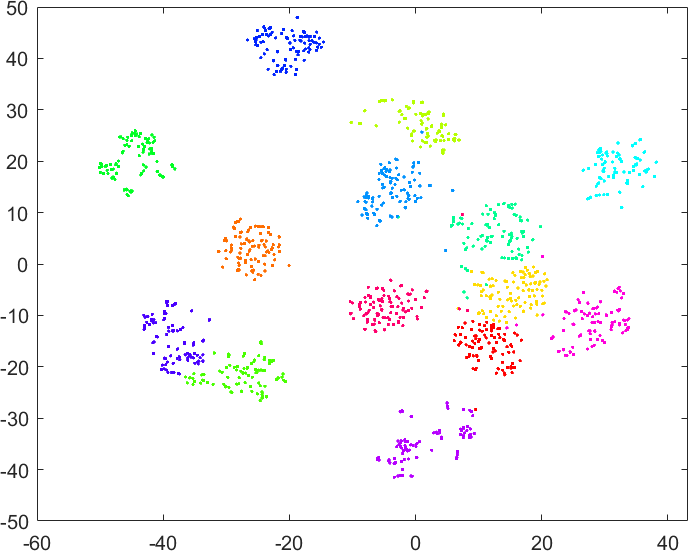}} &
\subfloat[$t$-SNE of sketches in shared latent space]{\includegraphics[width=0.47\linewidth]{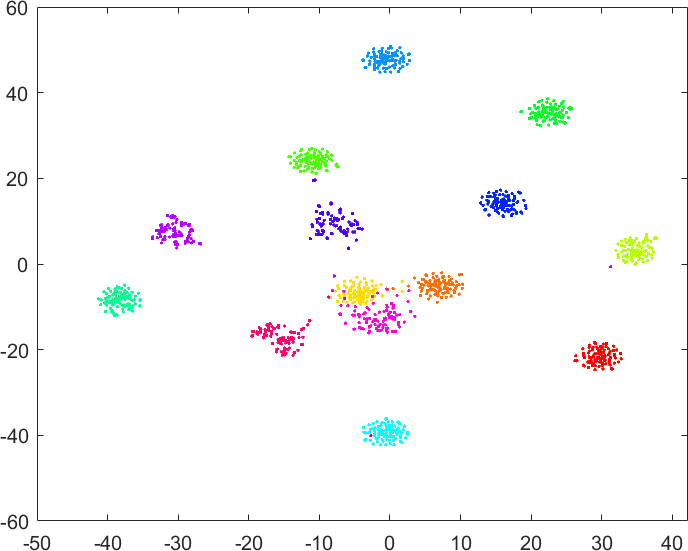}}
\end{tabular}
\caption{Two-dimensional scatter plots of high-dimensional features generated by $t$-SNE for the image and sketch datasets (a)--(b) before processing and (c)--(d) post-training in the shared latent space (trained with a fixed-semantic vector). The colours represent different classes in the dataset.}
\label{fig:tsne}
\end{figure*}

\newpage
\subsection{Retrieval results}
\begin{figure*}[!h] 
\centering 
\begin{tabular}{cccc}
\subfloat[Tennis]{{\includegraphics[height=1.5cm]{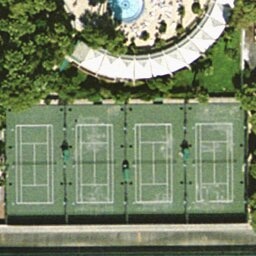}}} 
{\fcolorbox{green}{yellow}{\includegraphics[height=1.5cm]{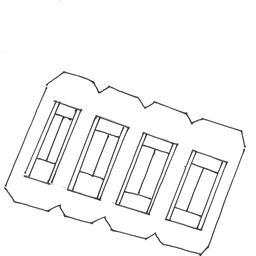}}}
{\fcolorbox{green}{yellow}{\includegraphics[height=1.5cm]{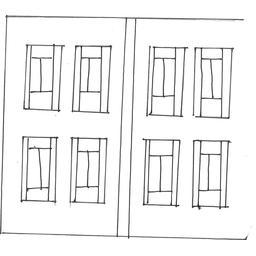}}}
{\fcolorbox{green}{yellow}{\includegraphics[height=1.5cm]{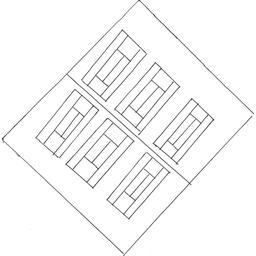}}}
{\fcolorbox{red}{yellow}{\includegraphics[height=1.5cm]{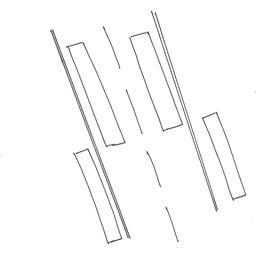}}}
{\fcolorbox{green}{yellow}{\includegraphics[height=1.5cm]{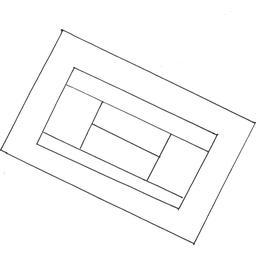}}}
{\fcolorbox{green}{yellow}{\includegraphics[height=1.5cm]{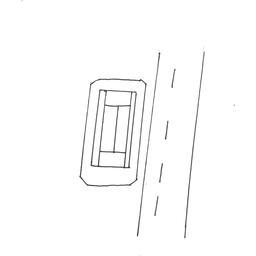}}}
{\fcolorbox{green}{yellow}{\includegraphics[height=1.5cm]{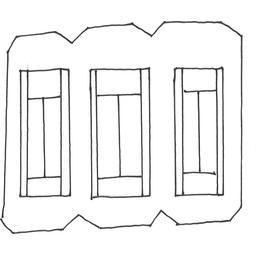}}}
{\fcolorbox{red}{yellow}{\includegraphics[height=1.5cm]{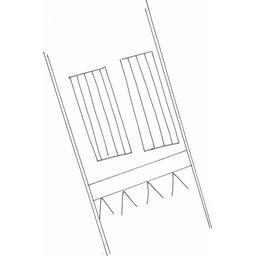}}}
{\fcolorbox{green}{yellow}{\includegraphics[height=1.5cm]{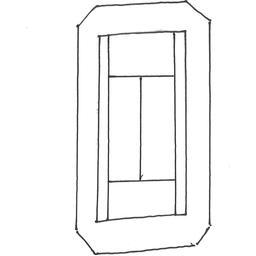}}}\\
\subfloat[Runway]{\includegraphics[height=1.5cm]{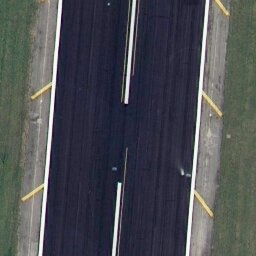}} 
{\fcolorbox{green}{yellow}{\includegraphics[height=1.5cm]{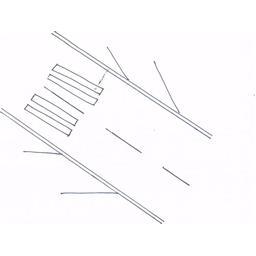}}}
{\fcolorbox{green}{yellow}{\includegraphics[height=1.5cm]{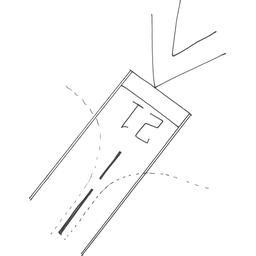}}}
{\fcolorbox{green}{yellow}{\includegraphics[height=1.5cm]{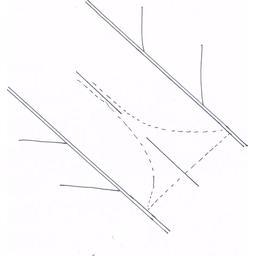}}}
{\fcolorbox{green}{yellow}{\includegraphics[height=1.5cm]{images2/55.jpg}}}
{\fcolorbox{red}{yellow}{\includegraphics[height=1.5cm]{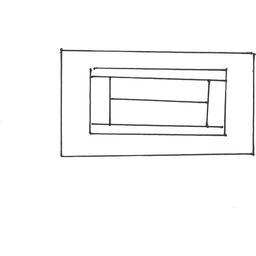}}}
{\fcolorbox{green}{yellow}{\includegraphics[height=1.5cm]{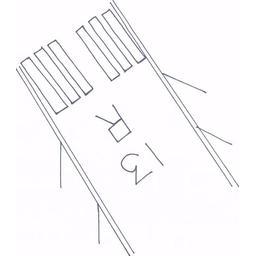}}}
{\fcolorbox{green}{yellow}{\includegraphics[height=1.5cm]{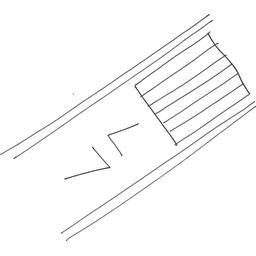}}}
{\fcolorbox{green}{yellow}{\includegraphics[height=1.5cm]{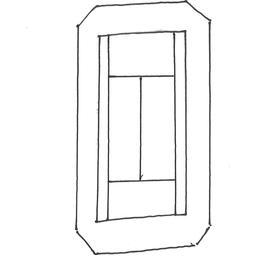}}}
{\fcolorbox{green}{yellow}{\includegraphics[height=1.5cm]{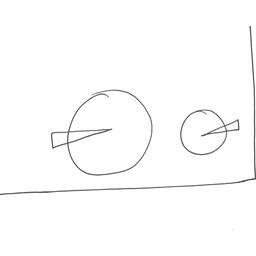}}}\\
\subfloat[Tank]{\includegraphics[height=1.5cm]{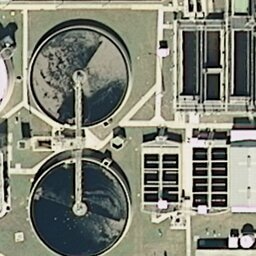}} 
{\fcolorbox{green}{yellow}{\includegraphics[height=1.5cm]{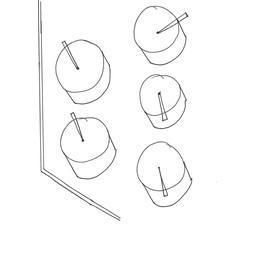}}}
{\fcolorbox{green}{yellow}{\includegraphics[height=1.5cm]{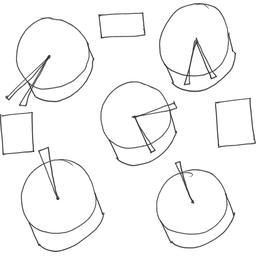}}}
{\fcolorbox{green}{yellow}{\includegraphics[height=1.5cm]{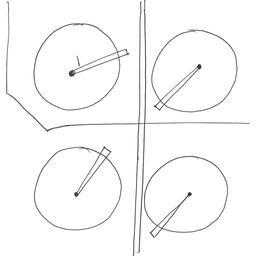}}}
{\fcolorbox{green}{yellow}{\includegraphics[height=1.5cm]{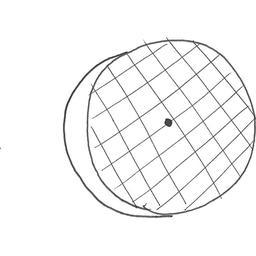}}}
{\fcolorbox{green}{yellow}{\includegraphics[height=1.5cm]{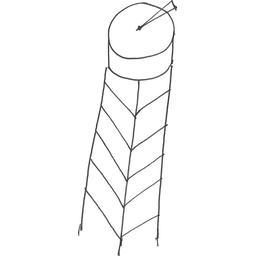}}}
{\fcolorbox{green}{yellow}{\includegraphics[height=1.5cm]{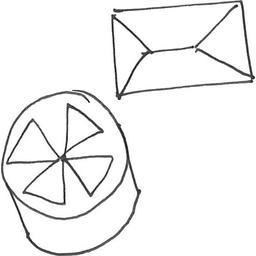}}}
{\fcolorbox{green}{yellow}{\includegraphics[height=1.5cm]{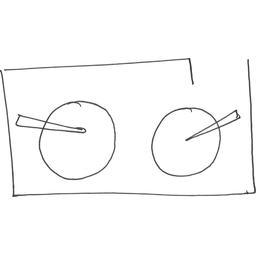}}}
{\fcolorbox{green}{yellow}{\includegraphics[height=1.5cm]{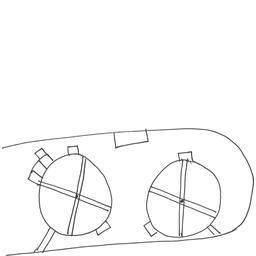}}}
{\fcolorbox{green}{yellow}{\includegraphics[height=1.5cm]{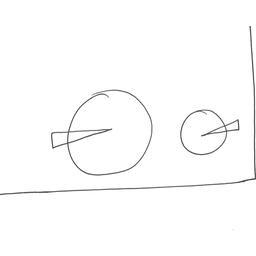}}}\\
\subfloat[River]{\includegraphics[height=1.5cm]{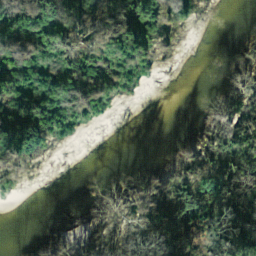}} 
\fcolorbox{green}{yellow}{\includegraphics[height=1.5cm]{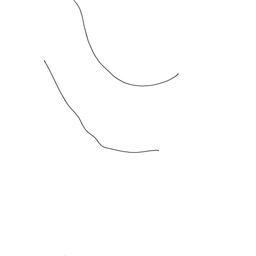}}
\fcolorbox{green}{yellow}{\includegraphics[height=1.5cm]{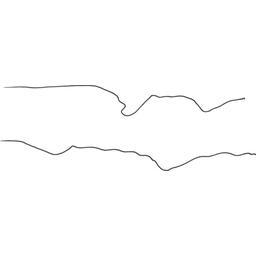}}
\fcolorbox{green}{yellow}{\includegraphics[height=1.5cm]{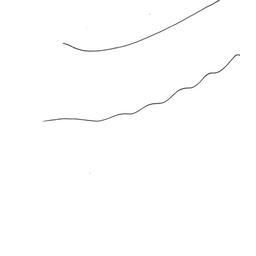}}
\fcolorbox{green}{yellow}{\includegraphics[height=1.5cm]{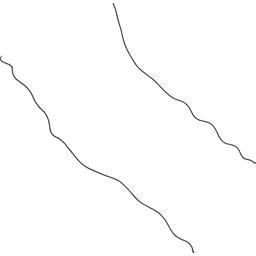}}
\fcolorbox{green}{yellow}{\includegraphics[height=1.5cm]{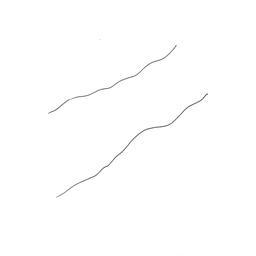}}
\fcolorbox{green}{yellow}{\includegraphics[height=1.5cm]{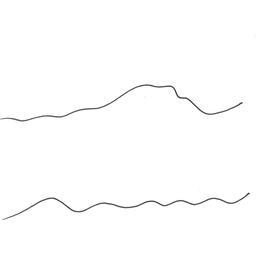}}
\fcolorbox{green}{yellow}{\includegraphics[height=1.5cm]{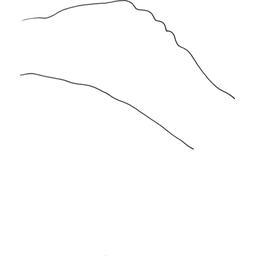}}
\fcolorbox{green}{yellow}{\includegraphics[height=1.5cm]{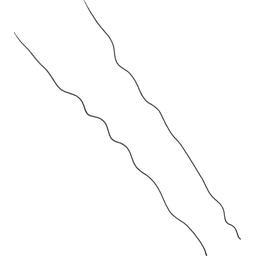}}
\fcolorbox{green}{yellow}{\includegraphics[height=1.5cm]{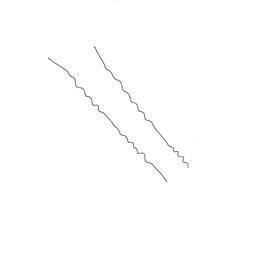}}\\
\subfloat[Runway]{\includegraphics[height=1.5cm]{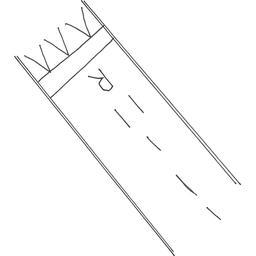}} 
\fcolorbox{green}{yellow}{\includegraphics[height=1.5cm]{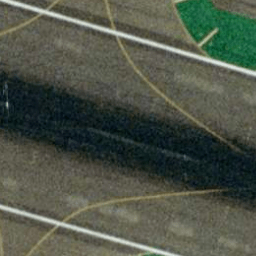}}
\fcolorbox{green}{yellow}{\includegraphics[height=1.5cm]{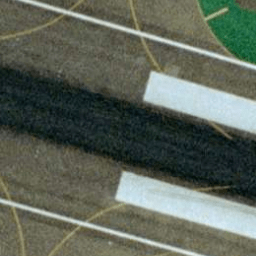}}
\fcolorbox{green}{yellow}{\includegraphics[height=1.5cm]{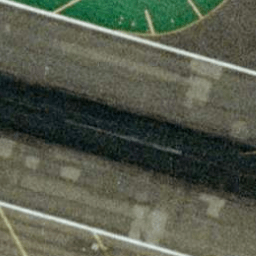}}
\fcolorbox{green}{yellow}{\includegraphics[height=1.5cm]{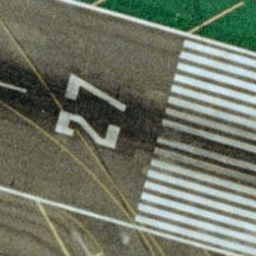}}
\fcolorbox{green}{yellow}{\includegraphics[height=1.5cm]{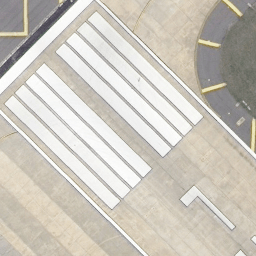}}
\fcolorbox{green}{yellow}{\includegraphics[height=1.5cm]{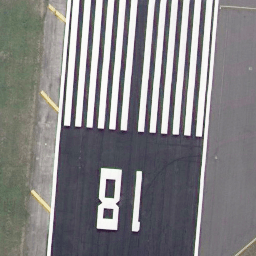}}
\fcolorbox{green}{yellow}{\includegraphics[height=1.5cm]{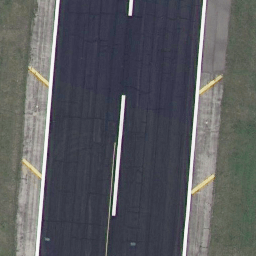}}
\fcolorbox{green}{yellow}{\includegraphics[height=1.5cm]{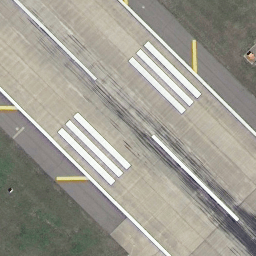}}
\fcolorbox{green}{yellow}{\includegraphics[height=1.5cm]{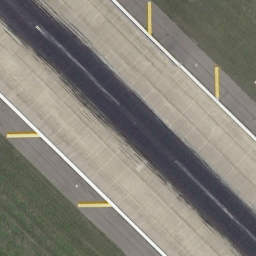}}\\
\subfloat[Tank]{\includegraphics[height=1.5cm]{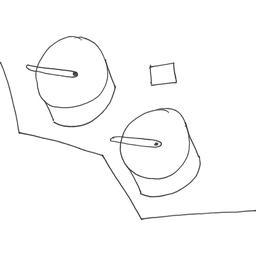}} 
\fcolorbox{green}{yellow}{\includegraphics[height=1.5cm]{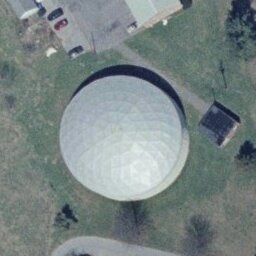}}
\fcolorbox{red}{yellow}{\includegraphics[height=1.5cm]{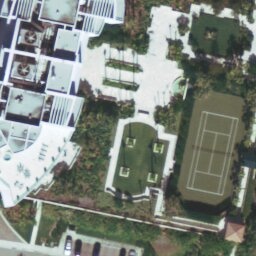}}
\fcolorbox{green}{yellow}{\includegraphics[height=1.5cm]{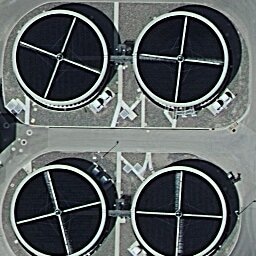}}
\fcolorbox{green}{yellow}{\includegraphics[height=1.5cm]{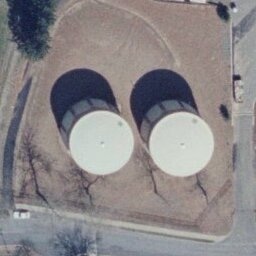}}
\fcolorbox{green}{yellow}{\includegraphics[height=1.5cm]{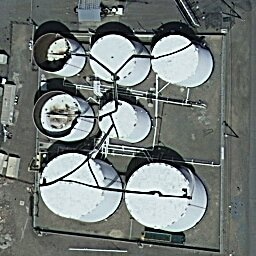}}
\fcolorbox{green}{yellow}{\includegraphics[height=1.5cm]{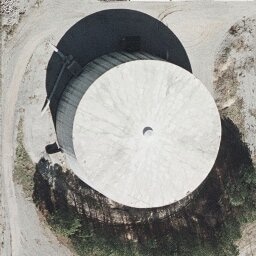}}
\fcolorbox{green}{yellow}{\includegraphics[height=1.5cm]{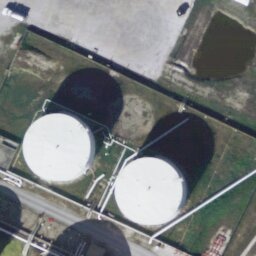}}
\fcolorbox{green}{yellow}{\includegraphics[height=1.5cm]{images2/storagetanks05.jpg}}
\fcolorbox{green}{yellow}{\includegraphics[height=1.5cm]{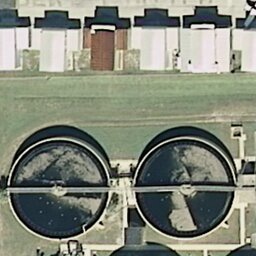}}\\
\subfloat[River]{\includegraphics[height=1.5cm]{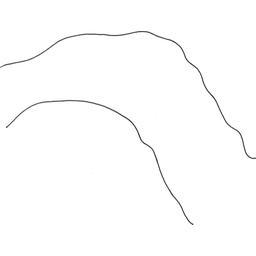}} 
\fcolorbox{green}{yellow}{\includegraphics[height=1.5cm]{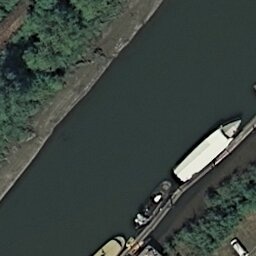}}
\fcolorbox{green}{yellow}{\includegraphics[height=1.5cm]{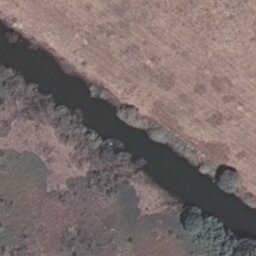}}
\fcolorbox{green}{yellow}{\includegraphics[height=1.5cm]{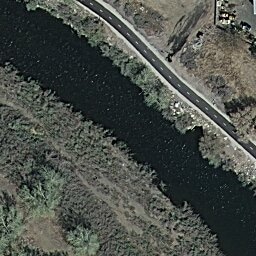}}
\fcolorbox{green}{yellow}{\includegraphics[height=1.5cm]{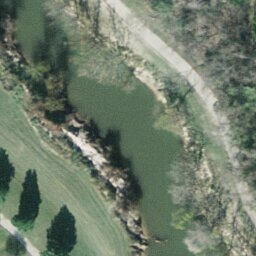}}
\fcolorbox{green}{yellow}{\includegraphics[height=1.5cm]{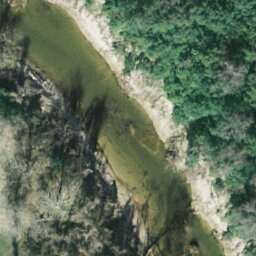}}
\fcolorbox{green}{yellow}{\includegraphics[height=1.5cm]{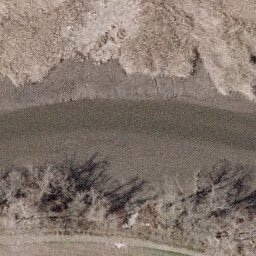}}
\fcolorbox{green}{yellow}{\includegraphics[height=1.5cm]{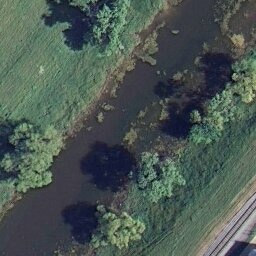}}
\fcolorbox{green}{yellow}{\includegraphics[height=1.5cm]{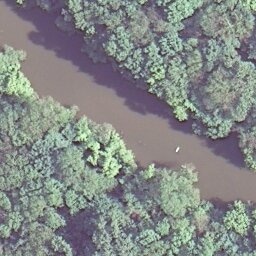}}
\fcolorbox{green}{yellow}{\includegraphics[height=1.5cm]{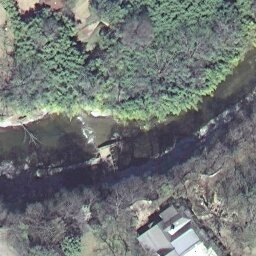}}\\
\subfloat[Tennis] {\includegraphics[height=1.5cm]{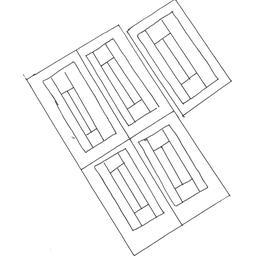}} 
\fcolorbox{green}{yellow}{\includegraphics[height=1.5cm]{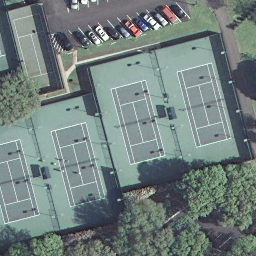}}
\fcolorbox{green}{yellow}{\includegraphics[height=1.5cm]{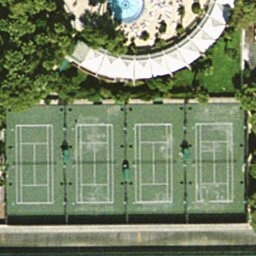}}
\fcolorbox{green}{yellow}{\includegraphics[height=1.5cm]{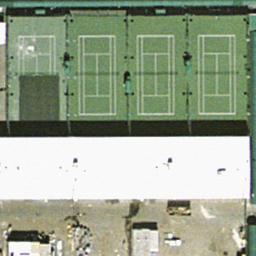}}
\fcolorbox{green}{yellow}{\includegraphics[height=1.5cm]{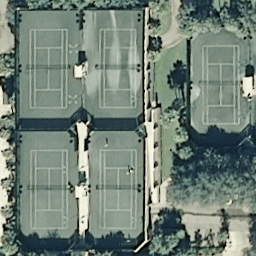}}
\fcolorbox{green}{yellow}{\includegraphics[height=1.5cm]{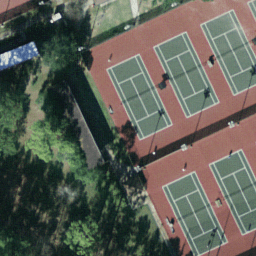}}
\fcolorbox{green}{yellow}{\includegraphics[height=1.5cm]{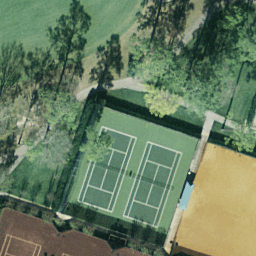}}
\fcolorbox{green}{yellow}{\includegraphics[height=1.5cm]{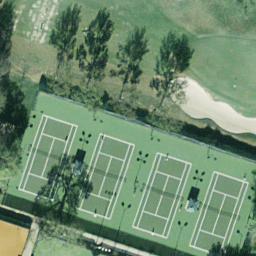}}
\fcolorbox{green}{yellow}{\includegraphics[height=1.5cm]{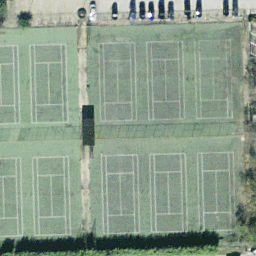}}
\fcolorbox{green}{yellow}{\includegraphics[height=1.5cm]{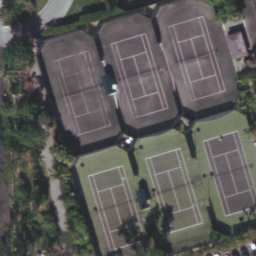}}\\
\end{tabular}
\caption{A few top retrieved results from the zero-shot cross-modal framework. The first four-row displays a few examples of \textbf{image$\rightarrow$sketch} retrievals, while the next four rows displays \textbf{sketch$\rightarrow$image} retrievals. The red outlines denote the falsely retrieved images, while the green contours denote the correctly retrieved image instances.}
\label{fig:retrieved}
\end{figure*}


%





\newpage
\section*{A Few Relevant References in Image Retrieval}


\subsection*{Uni-Modal Retrieval in Remote sensing}
\begin{enumerate}[label={[\arabic*]}]
\item  Michael Schroder, Hubert Rehrauer, Klaus Seidel, and Mihai Datcu. "Interactive learning and probabilistic retrieval in remote sensing image archives." \textit{IEEE Transactions on Geoscience and Remote Sensing} 38, no. 5 (2000): 2288-2298.

\item Marin Ferecatu, Michel Crucianu, and Nozha Boujemaa. "Retrieval of difficult image classes using svd-based relevance feedback." In \textit{Proceedings of the 6th ACM SIGMM international workshop on Multimedia information retrieval}, pp. 23-30. 2004.


\item Daniela Espinoza-Molina, and Mihai Datcu. "Earth-observation image retrieval based on content, semantics, and metadata." \textit{IEEE Transactions on Geoscience and Remote Sensing} 51, no. 11 (2013): 5145-5159.

\item Beg\"um Demir, and Lorenzo Bruzzone. "A novel active learning method in relevance feedback for content-based remote sensing image retrieval." \textit{IEEE Transactions on Geoscience and Remote Sensing} 53, no. 5 (2014): 2323-2334.

\item Beg\"um Demir, and Lorenzo Bruzzone. "Hashing-based scalable remote sensing image search and retrieval in large archives." \textit{IEEE Transactions on Geoscience and Remote Sensing} 54, no. 2 (2015): 892-904.

\item Bindita Chaudhuri, Beg\"um Demir, Lorenzo Bruzzone, and Subhasis Chaudhuri. "Region-based retrieval of remote sensing images using an unsupervised graph-theoretic approach."  \textit{IEEE Geoscience and Remote Sensing Letters} 13, no. 7 (2016): 987-991.

\item Bindita Chaudhuri, Beg\"um Demir, Subhasis Chaudhuri, and Lorenzo Bruzzone. "Multilabel remote sensing image retrieval using a semisupervised graph-theoretic method." \textit{IEEE Transactions on Geoscience and Remote Sensing} 56, no. 2 (2017): 1144-1158.

\item Yansheng Li, Yongjun Zhang, Xin Huang, Hu Zhu, and Jiayi Ma. "Large-scale remote sensing image retrieval by deep hashing neural networks." \textit{IEEE Transactions on Geoscience and Remote Sensing} 56, no. 2 (2017): 950-965.


\item Ushasi Chaudhuri, Biplab Banerjee, and Avik Bhattacharya. "Siamese graph convolutional network for content based remote sensing image retrieval." \textit{Computer Vision and Image Understanding} 184 (2019): 22-30.

\item Nagma Khan, Ushasi Chaudhuri, Biplab Banerjee, and Subhasis Chaudhuri. "Graph convolutional network for multi-label VHR remote sensing scene recognition." \textit{Neurocomputing} 357 (2019): 36-46.

\item Subhankar Roy, Enver Sangineto, Beg\"um Demir, and Nicu Sebe. "Metric-Learning-Based Deep Hashing Network for Content-Based Retrieval of Remote Sensing Images." \textit{IEEE Geoscience and Remote Sensing Letters} (2020).

\end{enumerate}

\subsection*{Cross-Modal retrieval in Remote sensing}
\begin{enumerate}[label={[\arabic*]}]
\setcounter{enumi}{11}
\item David Eigen, Christian Puhrsch, and Rob Fergus. "Depth map prediction from a single image using a multi-scale deep network." In \textit{Advances in Neural Information Processing Systems}, pp. 2366-2374. 2014.

\item Yansheng Li, Yongjun Zhang, Xin Huang, and Jiayi Ma. "Learning source-invariant deep hashing convolutional neural networks for cross-source remote sensing image retrieval." \textit{IEEE Transactions on Geoscience and Remote Sensing} 56, no. 11 (2018): 6521-6536.

\item Mao Guo, Chenghu Zhou, and Jiahang Liu. "Jointly Learning of Visual and Auditory: A New Approach for RS Image and Audio Cross-Modal Retrieval." \textit{IEEE Journal of Selected Topics in Applied Earth Observations and Remote Sensing} 12, no. 11 (2019): 4644-4654.

\item Ying Zhong, Wei Weng, Jianmin Li, and Shunzhi Zhu. "Collaborative Cross-Domain $k$ NN Search for Remote Sensing Image Processing." \textit{IEEE Geoscience and Remote Sensing Letters} 16, no. 11 (2019): 1801-1805.

\item Ushasi Chaudhuri, Biplab Banerjee, Avik Bhattacharya, and Mihai Datcu. "CMIR-Net: A deep learning based model for cross-modal retrieval in remote sensing." \textit{Pattern Recognition Letters} 131 (2020): 456-462.

\item Yaxiong Chen, Xiaoqiang Lu, and Shuai Wang. "Deep Cross-Modal Image-Voice Retrieval in Remote Sensing." \textit{IEEE Transactions on Geoscience and Remote Sensing} (2020).

\item Wei Xiong, Zhenyu Xiong, Yaqi Cui, and Yafei Lv. "A Discriminative Distillation Network for Cross-Source Remote Sensing Image Retrieval." \textit{IEEE Journal of Selected Topics in Applied Earth Observations and Remote Sensing} 13 (2020): 1234-1247.

\end{enumerate}
\subsection*{Zero-shot retrieval in Remote Sensing}
\begin{enumerate}[label={[\arabic*]}]
\setcounter{enumi}{18}

\item Gencer Sumbul, Ramazan Gokberk Cinbis, and Selim Aksoy. "Fine-grained object recognition and zero-shot learning in remote sensing imagery." \textit{IEEE Transactions on Geoscience and Remote Sensing} 56, no. 2 (2017): 770-779.

\item Aoxue Li, Zhiwu Lu, Liwei Wang, Tao Xiang, and Ji-Rong Wen. "Zero-shot scene classification for high spatial resolution remote sensing images." \textit{IEEE Transactions on Geoscience and Remote Sensing} 55, no. 7 (2017): 4157-4167.

\item Qian Song, Hui Chen, Feng Xu, and Tie Jun Cui. "EM Simulation-Aided Zero-Shot Learning for SAR Automatic Target Recognition." \textit{IEEE Geoscience and Remote Sensing Letters} 17, no. 6 (2019): 1092-1096.

\end{enumerate}

\subsection*{Sketch-based Image retrieval in Remote Sensing}
\begin{enumerate}[label={[\arabic*]}]
\setcounter{enumi}{21}
\item Tianbi Jiang, Gui-Song Xia, and Qikai Lu. "Sketch-based aerial image retrieval." \textit{IEEE International Conference on Image Processing} (ICIP), pp. 3690-3694. IEEE, 2017.

\item Fang Xu,  Ruixiang Zhang, Wen Yang, and Gui-Song Xia. "Mental Retrieval of Large-Scale Satellite Images Via Learned Sketch-Image Deep Features." \textit{IEEE International Geoscience and Remote Sensing Symposium}, pp. 3356-3359. IEEE, 2019.

\item Fang Xu,  Wen Yang, Tianbi Jiang, Shijie Lin, Hao Luo, and Gui-Song Xia. "Mental Retrieval of Remote Sensing Images via Adversarial Sketch-Image Feature Learning." \textit{IEEE Transactions on Geoscience and Remote Sensing} (2020).
\end{enumerate}


\subsection*{Review papers}
\begin{enumerate}[label={[\arabic*]}]
\setcounter{enumi}{24}

\item Xin-Yi Tong, Gui-Song Xia, Fan Hu, Yanfei Zhong, Mihai Datcu, and Liangpei Zhang. "Exploiting deep features for remote sensing image retrieval: A systematic investigation." \textit{IEEE Transactions on Big Data} (2019).

\item Peng Xu. "Deep learning for free-hand sketch: A survey." \textit{arXiv preprint} arXiv:2001.02600 (2020).

\end{enumerate}

\section*{Access to Dataset and Codes}
The Earth on Canvas dataset and the codes developed in this work are made available at: \\ \url{https://github.com/ushasi/Earth-on-Canvas-dataset-sample}.
\section*{Acknowledgements}
The authors want to thank Ms.\ Tanushree~Bhattacharya and Mr.\ Shivam~Pande for helping us creating the dataset and proposing its name.
